\documentclass[conference]{IEEEtran}
\IEEEoverridecommandlockouts
\usepackage{enumitem}
\usepackage{float}
\usepackage{capt-of}
\usepackage{booktabs}
\usepackage{amsmath,amssymb,amsfonts}
\usepackage{amsthm}
\usepackage{algorithmic}
\usepackage{graphicx}
\usepackage{textcomp}
\usepackage{xcolor}
\usepackage[british]{babel}   
\usepackage{microtype}
\usepackage{siunitx}
\usepackage{todonotes}

\floatstyle{plaintop}
\restylefloat{table}
\usepackage{tikz}
\usepackage{circuitikz}
\usetikzlibrary{backgrounds}
\usepackage{bbm}

\usetikzlibrary{arrows.meta,positioning,fit,calc,shapes,backgrounds}
\theoremstyle{definition}          
\newtheorem*{definition}{Definition}
\def\BibTeX{{\rm B\kern-.05em{\sc i\kern-.025em b}\kern-.08em
    T\kern-.1667em\lower.7ex\hbox{E}\kern-.125emX}}
\usepackage[hidelinks]{hyperref}
\begin{document}

\title{Quantifying Meaning with SCAPE-T: Semantic Complexity from Encoder-Only Transformer Attention\\}

\author{\IEEEauthorblockN{Maxim Khovansky}
\IEEEauthorblockA{\textit{EECS} \\
\textit{Queen Mary University of London}\\
London, United Kingdom}
\thanks{Corresponding author: maxkho00@gmail.com. This work was completed as part of the first author's final-year MSc project.}
\and
\IEEEauthorblockN{Lev Mukhanov}
\IEEEauthorblockA{
\textit{EECS} \\
\textit{Queen Mary University of London} \\
London, United Kingdom}
}

\maketitle

\begin{abstract}
Automated semantic complexity estimation remains underexplored in NLP, despite important applications in linguistics and interpretability. This paper defines semantic complexity as the code length required to uniquely specify a text’s semantic meaning, and operationalises it via SCAPE-T, an attention-derived metric based on encoder-only transformers. Existing measures either model reader effort (readability, lexical difficulty), which correlates with surface features (word length, syntax), or quantify surface features directly (type–token ratio, word frequency). SCAPE-T targets semantic integration directly by combining three submetrics -- (i) inverse-weighted self-focus (token→self), (ii) CLS focus (token→\texttt{[CLS]}), (iii) inter-head redundancy (cosine similarity across heads) -- using layer-weighted aggregation; data-driven learning is confined to hyperparameter tuning for calibration. Development uses philosophical prose vs MCTest stories; evaluation is on children’s stories vs technical Wikipedia articles, with pairwise accuracy on contrasts as the primary criterion. Baselines include readability, syntactic/lexical, and LLM-derived measures. SCAPE-T substantially outperforms all tested baselines under word-length control, with a pairwise accuracy of 78.5\% (95\% bootstrap CI: 76.1–80.4, N=10k pairwise comparisons; +12.7pp over strongest baseline); without that control, readability heuristics overtake, consistent with their reliance on surface features. On a short-terms abstractness dataset, SCAPE-T also leads. Ablations show each submetric contributes materially. Overall, the results show SCAPE-T isolates conceptual structure from lexical/syntactic confounds better than existing methods, and generalises across two held-out domains. The implementation is released as a Python package, installable via \texttt{pip install git+https://github.com/MaxKho/SCAPE.git}, with resources available at \url{https://github.com/MaxKho/SCAPE}.
\end{abstract}


\section{Introduction}
\label{sec:intro}

Automated semantic complexity scoring has clear utility across NLP, linguistics, and other domains, yet research into it remains scarce. A reliable measure could support: (i) empirical tests of the long-standing \emph{equi-complexity hypothesis} in linguistic typology \cite{Shcherbakova-2023, sampson2009complexity}; (ii) curriculum design and automated sequencing of instructional materials (from simpler to harder content) in education and assessment; and (iii) dataset construction, domain adaptation, and summarisation control, where complexity-aware sampling or targets are desirable. When operationalised via model-internal processing signals (as in this paper), analysing the metric's systematic limitations could also aid interpretability and model diagnostics -- for example, by flagging inputs that elicit a disproportionate semantic processing load, thereby localising where representation learning struggles with semantic integration.

Most work on language complexity focuses on readability or lexical sophistication, leaving semantic complexity underdefined. Even works aimed primarily at semantic complexity often adopt extensional characterisations such as abstractness, or informal notions such as “meaning density” \cite{siddharthan2006simplification}; such definitions are sufficient for ground-truth labelling, but are not rigorous enough to derive metrics from first principles. Consequently, this paper proposes the following definition:

\begin{definition}
The \emph{semantic complexity} of a text is the minimum per-token code length required to uniquely specify the text's semantic content relative to a fixed situation model\footnote{``Empirical world state’' here denotes a set of extra-linguistic facts sufficient to ground empirical lexical denotations (e.g. \emph{rock}, \emph{river}); it is ontic and agent-independent. ``Situation model’' denotes a specific semantic representation of said empirical world state.}\cite{Zwaan-2001} and empirical world state\footnotemark[\value{footnote}].
\end{definition}

As such, it can be conceptualised as the Kolmogorov complexity of semantic content normalised by token length and conditional on a fixed empirical world state and situation model. The definition is expected to be approximately model-invariant: admissible situation models that preserve semantic granularity should induce the same ranking up to a monotone transform\footnote{This assumption is to be tested in future work by applying SCAPE and SCAPE-T across different encoders (see \S\ref{sec:conclusion}).}. On this view, complexity increases as a text introduces, binds, or interrelates more propositions. Conversely, discrepancies in measured semantic complexity when different situation models are used can indicate that the used situation models are flawed (see the discussion of semantic integration blindspots in \S\ref{sec:analysis}).
Despite its utility, the literature offers few operationalisations that target this notion directly.

Existing measures tend to quantify reader effort or predictability rather than semantic content. Readability formulae and human-anchored lexical difficulty measures, such as Flesch–Kincaid, age-of-acquisition norms, frequency-based proxies, and ML models trained on human difficulty ratings, correlate strongly with surface features such as word length, frequency, and syntactic configuration \cite{flesch1948readability,kincaid1975readability,kuperman2012aoa,brysbaert2009subtlex,jarvis2010mtld}, confounding the semantic complexity signal this paper aims to isolate. Traditional semantic frameworks that depend on human domain labelling or handcrafted inventories do not scale to large cross-domain or cross-lingual corpora and are sensitive to the biases of the annotators and the chosen domain definitions \cite{pollard-biermann-2000-measure}. Alternative approaches include language-model perplexity and Minimum Description Length (MDL)-based approaches. Language-model perplexity measures the per-token cross-entropy under a language model trained on some corpus; while this approach avoids substantial dependence on surface features, it largely measures deviation from the model's training distribution (e.g., domain shift, rare lexico-syntactic patterns) rather than conceptual load. Minimum Description Length (MDL)-based approaches minimise the code length of the observed string under a model, so they compress on text symbols rather than on semantic meaning; their scores therefore still track surface features like word length/frequency and syntax more than semantic integration \cite{rissanen1978mdl}. Defining an MDL over meanings would require an explicit semantic representation and supervision, reintroducing human bias and limiting scalability. In short, prior work leaves us with strong tools for lexical and syntactic difficulty, but no generally adopted metric that directly targets semantic integration.

This paper addresses this gap with \textbf{SCAPE-T}: \emph{Semantic Complexity from Attention Patterns in Encoders} (SCAPE), explored here in its \emph{encoder-only Transformer} variant (T), which derives three complementary attention-derived submetrics directly from the specified definition (see \S\ref{sec:method} for further details): (i) \emph{inverse-weighted self-focus} (proportion of total attention from each token to itself) as a proxy for token interdependence; (ii) \emph{CLS focus} (proportion of total attention directed at the CLS token) as a proxy for emergent sentence-level meaning; and (iii) \emph{inter-head redundancy}, which penalises duplicated attention patterns via cross-head similarity. These submetrics are integrated via layer-weighted aggregation, which emphasises middle-to-late layers using a linear combination of uniform and Gaussian layer weights. For calibration and scaling, a set of hyperparameters -- namely the submetric exponents, a reciprocal-penalty scale, a head-similarity threshold for redundancy, and the layer-weighting parameters -- are tuned on token-length-matched development contrasts; no additional task supervision or model-weight training is performed to limit overfitting. In this paper, SCAPE-T is instantiated with BERT, but the approach is model-agnostic among encoders that expose attention and a sequence-level aggregator (e.g., modern BERT-family encoders and encoder–decoder models that retain a \texttt{[CLS]}-like token on the encoder side). In such cases, hyperparameters would likely require re-tuning.

Because semantic complexity ground truth is largely unavailable due to the ambiguity in existing definitions and absence of reliable metrics, proxies have to be relied on for evaluation. To minimise subjective bias, the proxies are chosen based on popular consensus: it is widely understood that dense philosophical prose is generally more semantically complex than simple descriptive statements, and that technical analysis of abstract concepts is generally more semantically complex than children's stories aimed at very young readers \cite{Graesser2003CoherenceRelations}. Accordingly, the development contrast pairs sentences scraped from Hegel and Kant with sentences from MCTest stories, and the test contrast pairs sentences from Wikipedia articles on technical topics with sentences from children’s stories for very young readers via Project Gutenberg; the domain and genre shift from philosophical prose to encyclopaedic content, and from MCTest to children’s stories, is introduced to test generalisation. Since the ground truth consists of binary contrasts, \emph{pairwise accuracy} is chosen as the primary evaluation metric: given one simple and one complex sentence, the metric should assign the higher score to the complex one. Additionally, a short-terms \emph{abstractness} set is generated with OpenAI o3 under physical-embodiment criteria: while still only a proxy, abstractness helps distinguish semantic complexity from lexical/syntactic complexity and probes generalisation in a style- and domain-neutral setting (one--two-word terms, no punctuation, no style or domain cues). Baselines include DHH lexicon complexity \cite{Alonzo-TSAR-2022}; MDL bits per byte compressor; GPT-2 perplexity; Flesch–Kincaid; Zipf frequency; BERT token length; type–token ratio; and mean dependency-tree depth. All metrics are computed on both the raw splits and a \emph{word-length-controlled} variant obtained by sampling so that the word-length distributions of the two sets closely match, as well as on the abstractness dataset. Under length control, SCAPE-T substantially outperforms all baselines on held-out domains (+12.7pp over the strongest baseline); without control, readability heuristics overtake, consistent with their dependence on surface features like word length. SCAPE-T also leads on the abstractness set, providing further evidence of generalisability.

\textbf{Contributions.} This paper (1) formalises \emph{semantic complexity} as the per-token description length relative to a fixed situation model and empirical world state; (2) introduces \textbf{SCAPE-T}, one of the first automated metrics to directly target \emph{semantic} (rather than lexical/syntactic/predictive) complexity; (3) proposes a novel operationalisation of language complexity based on internal processing (encoder attention structure) rather than data-driven classification or regression; (4) establishes an evaluation protocol that isolates semantics from lexical confounds via word-length control and pairwise accuracy on widely accepted proxy contrasts (Hegel/Kant vs.\ MCTest; technical Wikipedia vs.\ children’s stories), complemented by a style- and domain-neutral abstractness set; and (5) demonstrates that, under length control, SCAPE-T substantially outperforms strong readability, lexical, syntactic, LLM, and compressor baselines on held-out domains (+12.7pp over the strongest baseline), also leading on the abstractness set, with ablations confirming that each attention-derived submetric contributes.

\section{Related Work}

\subsection{Readability and Psycholinguistic Predictors}
Text complexity in NLP is most often operationalised through lexical and syntactic proxies. Readability formulae (e.g., Flesch–Kincaid) model reader effort via surface features such as sentence length, syllables per word, and word length \cite{flesch1948readability,kincaid1975readability,collins-thompson2014readability}. Feature-engineering approaches combine vocabulary statistics and structural cues: frequency lists and Zipf norms \cite{brysbaert2009subtlex}, type–token ratio for lexical diversity \cite{kettunen2014ttr}, part-of-speech and morphology, parse-based indicators such as mean dependency distance and tree depth \cite{vajjala2015syntactic,dellorletta2014readability}. Cohesion and discourse indices (e.g., Coh-Metrix) add connectives, referential overlap, and LSA-based coherence \cite{graesser2014cohmetrix,feng2010comparison,flor2013lingfeat,crossley2017tool}. These methods are effective predictors of readability and lexical difficulty, but they target surface predictability and form rather than semantic content.

Psycholinguistic features correlate with perceived difficulty and are widely used in complexity estimation. Age of acquisition, concreteness, polysemy, and familiarity provide human-anchored complexity information \cite{kuperman2012aoa,brysbaert2014concreteness,collins-thompson2014readability}. Such resources are informative about vocabulary difficulty, yet they embed human and genre biases and do not directly isolate semantic integration.

\subsection{Information-Theoretic, Formal, and Discourse-Based Proxies}
Distributional semantics has been used to approximate semantic content. Latent Semantic Analysis and related vector-space methods capture topical breadth and cohesion for readability and assessment \cite{landauer2011lsa,graesser2014cohmetrix,chersoni2016measuring}. Information-theoretic formulations quantify sequence predictability through language-model cross-entropy or compression. Minimum Description Length minimises model plus data code length \cite{rissanen1978mdl}; in practice, both perplexity and compressor bits-per-byte track token-level predictability and redundancy rather than the amount of information needed to specify semantic content. These measures are therefore sensitive to frequency, morphology, and syntax, not to semantic integration per se. This has been formalised in probing terms using MDL to assess representational predictability rather than semantic content \cite{pimentel2020mdl}.

Formal frameworks define semantic complexity relative to an explicitly modelled meaning space. Pollard and Biermann \cite{pollard-biermann-2000-measure} count the domain’s talkable entities, attributes, and relations to obtain an information-theoretic measure of domain breadth. The DAST model \cite{besharati2019dast} represents meanings in a lattice of “intuitions’’ and computes complexity from the lattice structure using a formal tuple. Both approaches provide principled semantics-grounded measures, but each presupposes a hand-crafted semantic representation or inventory and typically requires manual or rule-based construction. This limits automation and scalability to open-domain text.

Beyond sentence-level readability, discourse coherence models target how mentions and propositions are organised. Entity-grid and entity-graph methods \cite{barzilay2008entity,guinaudeau2013graph} capture transitions and connectivity that correlate with coherence and quality. RST-style discourse features and connectives have been shown to improve readability prediction over surface features alone \cite{pitler2008readability,feng2010comparison}. These lines more directly touch semantic organisation than syllable counts, but they still rely on discrete parsers and discourse inventories and are sensitive to genre and parser error. 

\subsection{Attention- and Representation-Based Methods}
Neural methods have begun to target complexity via model-internal processing. Chen et~al. \cite{chen2025autoencoder} propose a bidirectional attention-based autoencoder for \emph{semantic complexity} grading. The model reports strong classification accuracy, and the design suggests that reconstruction difficulty could reflect conceptual load. However, the system is trained on, and evaluated against, \emph{lexical complexity} annotations (SemEval-style labels) and includes vocabulary-complexity features, so it primarily captures surface difficulty rather than semantic integration. Autoencoder reconstruction loss remains a promising alternative proxy if decoupled from lexical supervision and hand-crafted surface features.

Analyses of transformer encoders show that attention heads capture syntactic and semantic relations at different layers \cite{clark2019bert,tenney-etal-2019-bert,rogers2020primer}. Head redundancy and pruning studies indicate overlapping patterns across heads \cite{michel2019sixteen}. These findings motivate complexity metrics that read encoder-internal structure rather than model outputs. SCAPE-T follows this line by aggregating attention-derived signals, including head redundancy, to target semantic integration, in contrast to output-likelihood measures and surface-feature heuristics. These findings directly motivate penalties for cross-head similarity \cite{voita2019analyze,kovaleva2019bert}.

Despite the aforementioned attention-based methods, the explanatory power of attention remains contested. Adversarial and deletion analyses question whether raw attention weights explain model predictions \cite{jain2019attention,serrano2019attention}, while counter-arguments show that, under appropriate desiderata, attention can be explanatory or at least diagnostic \cite{wiegreffe2019attention}. Multi-hop attributions such as attention rollout/flow aggregate contributions across layers and paths, addressing some locality issues \cite{abnar2020rollout}. Gradient- and relevance-based alternatives (e.g., gradient$\times$input, integrated gradients \cite{Sundararajan2017IG}, and transformer-specific relevance propagation) provide complementary internal signals that do not rely on raw attention magnitudes \cite{chefer2021transformer}. SCAPE-T's use of attention therefore should be read as a diagnostic measure of integration, not as a causal explanation for model predictions.

SemEval-2021 Task~1 provides lexical complexity labels for words and expressions in context \cite{shardlow-etal-2021-semeval}. Such datasets are valuable for vocabulary difficulty but do not constitute ground truth for semantic complexity. The 10k Prompts Ranked dataset \cite{DIBT2023} captures LLM prompt difficulty as judged by users. These rankings primarily reflect the difficulty of completing the specified task, not the intrinsic semantic complexity of the prompt text. Neither resource directly evaluates a metric that aims to isolate conceptual structure from lexical and syntactic confounds.

\subsection{Summary}

The literature offers (i) strong surface and psycholinguistic predictors of readability, (ii) predictability and compression proxies sensitive to lexical and domain factors, (iii) rigorous but non-automated semantic frameworks, (iv) discourse-level models of entity and relation organisation, and (v) initial neural methods trained on lexical labels. What is still missing is a scalable, label-free metric that targets semantic integration by leveraging model-internal evidence. This gap motivates the approach introduced in this paper.

\section{Methodology}
\label{sec:method}

\subsection{Why encoder self-attention?}
This paper aims to measure the semantic complexity of the \emph{input} sequence. Decoder self-attention in autoregressive models is conditioned on previously generated outputs, and encoder--decoder cross-attention explicitly conditions on output-side queries \cite{vaswani2017attention}. Both inject information about the \emph{output} hypothesis space into attention patterns, confounding a metric that characterises input text only. Encoder-only self-attention, by contrast, is conditioned solely on the input tokens, making it the appropriate substrate for an input-referential metric. BERT Base Uncased (evaluation mode) was chosen as the transformer model to reduce computational overhead and maximise scalability.

\subsection{Applying the definition to encoder-only transformers}
\label{sec:def-to-encoders}
This paper estimates the (input-referential) minimum code length required to specify a sequence's semantic content by reading signals from a \emph{frozen} encoder-only Transformer. The situation model is operationalised as the entire frozen encoder\footnote{There is evidence for representational convergence across modern encoders and emergent cross-lingual alignment in multilingual models \cite{wu-etal-2020-similarity,phang-etal-2021-finetuned,pires-etal-2019-multilingual,wu-dredze-2019-surprising}, suggesting -- but not guaranteeing -- that the situation model may be relatively stable under encoder swaps (including multilingual encoders). This paper treats cross-encoder/cross-lingual portability as a \emph{hypothesis} to be tested in future work, not  as an assumption; see \S\ref{sec:limitations}.} and understood as fixed and task-agnostic in the context of this paper: pretraining equips it with priors over semantic regularities, represented at least partially by attention distributions as well as token embeddings. The use of the entire frozen encoder (not embeddings alone) reflects the empirical fact that semantic information is largely carried by contextualised layers rather than static embeddings \cite{ethayarajh-2019-contextual,tenney-etal-2019-bert,jawahar-etal-2019-bert}. The empirical world state is operationalised as the model’s empirical parametric knowledge, which serves as a fixed general proxy for the empirical world state in the definition. This proxy is treated as adequately world-representative at everyday granularity given the large and diverse pretraining corpora of transformers such as BERT; additionally, transformer-based language models have been demonstrated to be capable knowledge bases \cite{petroni-etal-2019-language}. Nonetheless, notable coverage gaps remain, as discussed in \S\ref{sec:limitations}.

\textbf{Operational assumptions.}
The submetrics constituting SCAPE-T rest on three monotonicity assumptions linking latent code length to specific principles applicable to observable attention structure; each is justified as a chain of implications from the definition given in \S\ref{sec:intro}. For clarity, consider three latent loci of semantic load under a fixed situation model: (i) \emph{sentence-level emergence} (how much of the intended content exists only when the sentence is interpreted as a whole, beyond the resolved meanings of its constituents), (ii) \emph{subsentential emergence} (how much of the intended content exists only when two or more tokens are interpreted as a whole, beyond individual token content), and (iii) \emph{channel utilisation} (how many nonredundant relational channels\footnote{``Relational channel'' here denotes a processing pathway that represents relations among tokens. In SCAPE-T, relational channels are operationalised as attention heads (see \S\ref{sec:def-to-encoders}); channel independence refers to nonredundant semantic information, not merely distinct heads or attention patterns.} are engaged to encode the same input). Based on these loci, the following monotonicity assumptions are posited:

\begin{enumerate}[label=(A\arabic*), leftmargin=2.2em]
\item \textbf{Sentence-level emergence $\Rightarrow$ higher semantic complexity.}
Controlling for all other factors, if more meaning is emergent at the sentence level beyond the resolved meanings of its constituents, then more global relational content must be specified in addition to that constituent content; therefore, the minimum description needed to uniquely specify the sentence's semantic content is strictly larger, hence higher semantic complexity.

\item \textbf{Subsentential emergence $\Rightarrow$ higher semantic complexity.}
Controlling for all other factors, if more meaning is emergent within subsentential constituents beyond individual token content, then more relational content among tokens must be specified in addition to individual token content; therefore, the minimum description is strictly larger, hence higher semantic complexity.

\item \textbf{Channel utilisation $\Rightarrow$ higher semantic complexity.}
Controlling for all other factors, if more relational channels contribute nonredundant semantic information in addition to that already encoded, then more information must be specified in the description; therefore, the minimum description is strictly larger, hence higher semantic complexity.
\end{enumerate}

The next subsection instantiates (A1)–(A3) using attention-derived submetrics from a frozen encoder: CLS focus as an estimator of sentence-level consolidation, inverse-weighted self-focus as an estimator of token interdependence, and head redundancy as a penalty for lower independent channel utilisation; it further shows that each is a monotone correlate of the underlying code length under these assumptions. These submetrics are intended as theoretically motivated \emph{correlates} rather than exhaustive estimators; SCAPE-T is therefore extensible and can incorporate additional correlates that admit analogous monotonic justifications.

\textbf{Notation.}
Let the tokenised input have $T$ positions after prefixing \texttt{[CLS]} at index $0$; let $\mathcal{I}=\{1,\dots,T-1\}$. For layer $\ell\!\in\!\{0,\dots,L\}$ and head $h\!\in\!\{1,\dots,H\}$, $A^{(\ell,h)}\!\in\![0,1]^{T\times T}$ is the row-stochastic attention matrix. $\epsilon$ is a small stabilising constant.

\subsubsection{Estimator 1. CLS focus (from A1)}
In BERT-style encoders, \texttt{[CLS]} aggregates sentence-level information \cite{devlin-etal-2019-bert}. Therefore, when sentence-level emergence increases, attention to \texttt{[CLS]} is expected to increase, while attention mass among non-\texttt{[CLS]} columns is expected to decrease. A scale-stable estimator is the ratio of the mean value of the former and the mean value of the latter:
\begin{equation}
\label{eq:cf}
\mathrm{CF}_{\ell,h}
=
(T-1)\frac{\sum_{j=0}^{T-1} A^{(\ell,h)}_{j,0}}
     {\sum_{j=0}^{T-1}\sum_{k\in\mathcal{I}} A^{(\ell,h)}_{j,k} \;+\; \epsilon}
\end{equation}
By (A1), higher $\mathrm{CF}_{\ell,h}$ monotonically indicates more sentence-level content that must be encoded, hence higher semantic complexity.

\subsubsection{Estimator 2 (pre-weighting). Self-Focus (from A2)}
For token $i$, greater contextual demand implies lower $A^{(\ell,h)}_{i,i}$ and more mass on other columns. Self-focus is then defined as the ratio of the mean self-directed attention over all non-\texttt{[CLS]} tokens and the global mean non-self-directed attention:
\begin{equation}
\label{eq:sf-raw-one}
\mathrm{SF}_{\ell,h}
\;=\;
T \frac{\;\sum_{i\in\mathcal{I}} A^{(\ell,h)}_{i,i}}
     {\,T - \sum_{j=0}^{T-1} A^{(\ell,h)}_{j,j}\;+\; \epsilon}.
\end{equation}
By (A2), lower $\mathrm{SF}_{\ell,h}$ monotonically indicates that more subsentential relational constraints must be specified to fix meaning, hence higher semantic complexity. The inverse weighting as well as scaling are introduced later in \S\ref{sec:composite}.

\subsubsection{Estimator 3. Head redundancy (from A3)}
Let $\tilde{A}^{(\ell,h)}$ be the attention map with the diagonal and the \texttt{[CLS]} column zeroed (so redundancy captures information not already measured by self/CLS links). L2-normalised flattened vectors are formed as
\[
v^{(\ell,h)} \;=\; \frac{\mathrm{vec}\!\big(\tilde{A}^{(\ell,h)}\big)}{\left\|\mathrm{vec}\!\big(\tilde{A}^{(\ell,h)}\big)\right\|_2+\epsilon}.
\]
For head $(\ell,h)$, the comparison set comprises all heads in strictly previous layers and the earlier-indexed heads in the same layer:
\[
\mathcal{P}_{\ell,h} \;=\; \{(\ell',h'):\,\ell'<\ell\}\,\cup\,\{(\ell,j):\,j<h\}.
\]
The maximum cosine similarity to prior heads is
\begin{equation}
\label{eq:sim}
\mathrm{sim}^{\star}_{\ell,h} \;=\;
\begin{cases}
\max\limits_{(\ell',h')\in\mathcal{P}_{\ell,h}} \; v^{(\ell,h)}\!\cdot v^{(\ell',h')}, & \mathcal{P}_{\ell,h}\neq\emptyset,\\[4pt]
0, & \mathcal{P}_{\ell,h}=\emptyset.
\end{cases}
\end{equation}
Note that, due to attention weights being nonnegative, the shared input, and chance overlap, attention heads are likely to have a baseline level of cosine similarity; to isolate true functional redundancy from these confounding variables, therefore, scores below a certain threshold need to be discounted.
With threshold $\tau\in[0,1]$, the redundancy coefficient is
\begin{equation}
\label{eq:r}
\mathrm{R}_{\ell,h} \;=\;
\begin{cases}
1, & \text{if } \mathrm{sim}^{\star}_{\ell,h}\le\tau,\\
1-\mathrm{sim}^{\star}_{\ell,h}, & \text{otherwise.}
\end{cases}
\end{equation}
This order definition (by head index within a layer, all heads from earlier layers) makes the estimator order-invariant to any permutation of later heads.
By (A3), larger $\mathrm{R}_{\ell,h}$ monotonically indicates more independent relational channels engaged, hence higher semantic complexity.

\subsubsection{Layer aggregation}
Evidence suggests a coarse progression across encoder layers: early layers emphasise lexical/syntactic processing, mid-to-late layers encode semantics most strongly, and the very top layers are most task-specific \cite{tenney-etal-2019-bert,jawahar-etal-2019-bert,ethayarajh-2019-contextual}. These findings motivate a roughly bell-shaped weighting profile that emphasises intermediate-to-late layers. Accordingly, this paper aggregates layers using a normalised mixture of a Gaussian kernel (centred at $\mu$ with variance $\sigma^2$) and uniform weights, with the uniform component allowing the strength of this emphasis to be calibrated:
\begin{equation}
\label{eq:weights}
W_{\ell} \;=\; (1-\eta)\,\frac{1}{L+1} \;+\; \eta\,
\frac{e^{\big(-\frac{(\ell-\mu)^2}{2\sigma^2}\big)}}{\sum_{l=0}^{L}e^{\big(-\frac{(\,l-\mu)^2}{2\sigma^2}\big)}},\quad \sum_{\ell=0}^{L} W_{\ell}=1.
\end{equation}

\subsubsection{Composite estimator}
\label{sec:composite}
The final score integrates all the submetrics specified above into a single composite measure: 

\begin{equation}
\label{eq:composite}
S = \mathrm{Base}^\gamma \cdot \mathrm{ISF}^\delta ,
\end{equation}

\noindent where:
\begin{itemize}
    \item \(\mathrm{Base}\) is computed as:
        \begin{equation}
        \label{eq:base}
        \mathrm{Base} =
        \sum_{\ell=0}^{L} W_{\ell}\Big(\tfrac{1}{H}\sum_{h=1}^{H} \mathrm{R}_{\ell,h}^{\alpha}\cdot(1+\mathrm{CF}_{\ell,h})^{\beta}\cdot(T-1)^{-\beta\zeta}\Big).
        \end{equation}
        (The length normalisation factor $(T-1)^{-\beta\zeta}$ compensates for the potential sequence length dependence introduced by the explicit $(T-1)$ factor in the CLS focus estimator (see (\ref{eq:cf})); the exponent $\zeta$ is retained to allow tuning on development data to accommodate more complex length dependencies in the underlying attention distributions.)
        \item \(\mathrm{ISF}\) is the Inverse-weighted Self-Focus, computed as:
         \begin{equation}
        \label{eq:isf}
        \mathrm{ISF}
        \;=\;
        1 \;+\; \frac{\kappa}{\,\displaystyle\sum_{\ell=0}^{L} W_{\ell}\Big(\tfrac{1}{H}\sum_{h=1}^{H}\mathrm{SF}_{\ell,h}\Big)\;+\; \epsilon}.
        \end{equation}
        (This term is applied outside of \(\mathrm{Base}\) because, empirically on the development set, this separation improves performance.)
        \item The remaining quantities are hyperparameters, detailed
        in \emph{Table \ref{tab:hyperparams}}:
\end{itemize}

\begin{table}[!b]
\centering
\caption{Hyperparameters.}
\vspace{6pt}
\label{tab:hyperparams}
\renewcommand{\arraystretch}{1.1}
\begin{tabular}{ll}
\toprule
\textbf{Name} & \textbf{Role} \\
\midrule
$\alpha$ & Exponent on redundancy $\mathrm{R}_{\ell,h}$ in (\ref{eq:base}) \\
$\beta$  & Exponent on CLS focus $\mathrm{CF}_{\ell,h}$ in (\ref{eq:base}) \\
$\gamma$ & Exponent on $\mathrm{Base}$ in (\ref{eq:composite}) \\
$\delta$ & Exponent on $\mathrm{ISF}$ in (\ref{eq:composite}) \\
$\zeta$  & Length normalisation exponent in (\ref{eq:base}) \\
$\kappa$ & Inverse-weighting scale in (\ref{eq:isf}) \\
$\tau$   & Similarity threshold for redundancy in (\ref{eq:r}) \\
$\eta$   & Gaussian layer weighting strength in (\ref{eq:weights}) \\
$\mu,\sigma$ & Mean and std.\ of Gaussian layer weighting in (\ref{eq:weights}) \\
\bottomrule
\end{tabular}
\end{table}

The input text is preprocessed by splitting on \texttt{\{.!?\}}, boilerplate cleaning, punctuation removal, basic stopword removal, and a few additional minor operations. The full preprocessing pipeline is detailed in Appendix \ref{app:preproc}. 

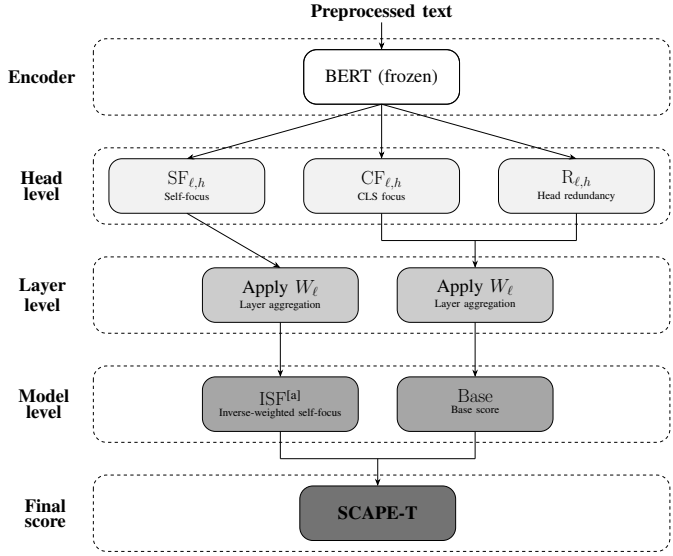
\begin{figure}[!t]
\centering
\resizebox{\columnwidth}{!}{%
\begin{circuitikz}[x=1cm, y=0.70cm]   
\tikzstyle{every node}=[font=\LARGE, align=center]
\tikzset{levelbox/.style={line width=0.6pt, rounded corners=12, draw=black}}
\colorlet{encfill}{white}  
\colorlet{headfill}{black!5}
\colorlet{layerfill}{black!20}
\colorlet{modfill}{black!35}
\colorlet{finalfill}{black!55}

\draw[levelbox, fill=encfill] (7.50,19.75) rectangle (12.50,17.25);

\draw[levelbox, fill=headfill] (1.25,14.75) rectangle (6.25,12.25);
\draw[levelbox, fill=headfill] (7.50,14.75) rectangle (12.50,12.25);
\draw[levelbox, fill=headfill] (13.75,14.75) rectangle (18.75,12.25);

\node[
  levelbox, fill=layerfill,
  anchor=center,
  minimum width=5.0cm,
  minimum height=1.75cm,
  inner ysep=0pt,
  align=center
] at (6.75,8.50)  
{\shortstack{Apply $W_{\ell}$\\ \normalsize Layer aggregation}};
\draw[levelbox, fill=layerfill] (10.50,9.75) rectangle (15.50,7.25);

\draw[levelbox, fill=modfill] (4.25,4.75)  rectangle (9.25,2.25);   
\draw[levelbox, fill=modfill] (10.50,4.75) rectangle (15.50,2.25);

\draw[levelbox, fill=finalfill] (7.375,-0.25) rectangle (12.375,-2.75); 

\draw [ line width=0.6pt , rounded corners = 12.0, dashed] (0.75,15.25) rectangle  (19.25,11.75);
\draw [ line width=0.6pt , rounded corners = 12.0, dashed] (0.75,10.25) rectangle  (19.25,6.75);
\draw [ line width=0.6pt , rounded corners = 12.0, dashed] (0.75,5.25) rectangle  (19.25,1.75);
\draw [ line width=0.6pt , rounded corners = 12.0, dashed] (0.75,0.25) rectangle  (19.25,-3.25);

\draw [line width=0.8pt, short] (10,12.25) -- (10,11);
\draw [line width=0.8pt, short] (16.25,12.25) -- (16.25,11);
\draw [line width=0.8pt, short] (10,11) -- (16.25,11);
\draw [line width=0.8pt, ->, >=Stealth] (13,11) -- (13,9.75);   

\draw [line width=0.8pt, ->, >=Stealth] (3.75,12.25) -- (6.75,9.75);   

\draw [line width=0.8pt, ->, >=Stealth] (6.75,7.25) -- (6.75,4.75);     

\draw [line width=0.8pt, ->, >=Stealth] (13,7.25) -- (13,4.75);

\draw [line width=0.8pt, short] (6.75,2.25) -- (6.75,1);  
\draw [line width=0.8pt, short] (13,2.25) -- (13,1);
\draw [line width=0.8pt, short] (6.75,1) -- (13,1);       

\draw [line width=0.8pt, ->, >=Stealth] (9.875,1) -- (9.875,-0.25);  

\draw [ line width=0.6pt , rounded corners = 12.0, dashed] (0.75,20.25) rectangle (19.25,16.75);

\node at (10.00,21.45) {\textbf{Preprocessed text}};
\draw [line width=0.8pt, ->, >=Stealth] (10.00,21.00) -- (10.00,19.75);
\draw [ line width=0.6pt , rounded corners = 12.0] (7.50,19.75) rectangle (12.50,17.25);
\node at (10.00,18.50) {BERT (frozen)};

\draw [line width=0.8pt, ->, >=Stealth] (10.00,17.25) -- (3.75,14.75);
\draw [line width=0.8pt, ->, >=Stealth] (10.00,17.25) -- (10.00,14.75);
\draw [line width=0.8pt, ->, >=Stealth] (10.00,17.25) -- (16.25,14.75);

\node[anchor=east] at (0.3,18.50) {\textbf{Encoder}};
\node[anchor=east] at (-0.10,13.50) {\shortstack{\textbf{Head}\\\textbf{level}}};
\node[anchor=east] at (0.00, 8.50) {\shortstack{\textbf{Layer}\\\textbf{level}}};
\node[anchor=east] at (0.10, 3.50) {\shortstack{\textbf{Model}\\\textbf{level}}};
\node[anchor=east] at (0.00,-1.50) {\shortstack{\textbf{Final}\\\\\textbf{score}}};

\node at (3.75,13.50) {\shortstack{$\mathrm{SF}_{\ell,h}$ \\ \normalsize Self-focus}};
\node at (10.00,13.50) {\shortstack{$\mathrm{CF}_{\ell,h}$ \\ \normalsize CLS focus}};
\node at (16.25,13.50) {\shortstack{$\mathrm{R}_{\ell,h}$ \\ \normalsize Head redundancy}};
\node[align=center, text height=5.2ex, text depth=.4ex] at (13.00,8.50) {\shortstack{Apply $\displaystyle W_{\ell}$ \\ \normalsize Layer aggregation}};
\node at (6.75,3.50)  {\shortstack{$\mathrm{ISF}$\textsuperscript{\text{[a]}} \\ \normalsize Inverse-weighted self-focus}};
\node[align=center, text height=4.0ex, text depth=.4ex] at (13.00,3.50) {\shortstack{$\mathrm{Base}$ \\ \normalsize Base score}};
\node at (9.875,-1.50) {\textbf{SCAPE-T}};

\end{circuitikz}
}%
\caption{High-level design of SCAPE-T with BERT. 
\\ {[a]} The post-aggregation transform $1+\kappa/(\cdot+\epsilon)$ is applied to obtain $\mathrm{ISF}$; see (\ref{eq:isf}).}
\label{fig:HLD}
\end{figure}

Scores are computed \emph{per sentence}; multi-sentence texts are assigned the arithmetic mean of their per-sentence scores. 
The final computation pipeline is detailed in \emph{Figure \ref{fig:HLD}}.

\subsection{Scope, correlates, and extensibility}
\label{subsec:scope}
The three base submetrics are \emph{theoretically motivated correlates} of the minimum code length under a fixed situation model; they are not claimed to exhaustively estimate code length. SCAPE-T is therefore extensible: additional correlates (e.g., other forms of cross-token integration or alternative capacity-use indicators) can be incorporated multiplicatively or via calibrated transforms, provided they satisfy a justified monotonic relationship with minimum code length and remain input-referential.

\section{Data \& Evaluation}
\subsection{Sentence datasets.}
Both sentence datasets were generated by producing simple vs complex contrasts. 

For the development complex group, 4,529 sentences from Hegel's \emph{Science of Logic} and 5,034 sentences from Kant's \emph{The Critique of Pure Reason} were scraped from Project Gutenberg, with headers and footers stripped, whitespace normalised. For the development simple group, 31977 sentences were scraped from MCTest-500 stories. The development set is \emph{token-length-controlled} via greedy matching on BERT token length to $n\!=\!200$ pairs.

For the test simple group, 143 sentences from \emph{Three Little Pigs} (\#18155), \emph{The Gingerbread Man} (\#25877), and the \emph{Children’s Hour} collection (\#11592) were scraped from Project Gutenberg, with Gutenberg boilerplate removed, bracketed tags stripped, and sentinel lines dropped (3+ uppercase letters, ``//'’, or ``EBook''). For the test complex group, 1123 sentences were scraped from Wikipedia pages on Artificial Intelligence, Quantum Field Theory, and Philosophy of Mind after citation/number artifact cleanup. A \emph{word-length-controlled} test set ($n\!=\!100$) -- \emph{WLC test} -- is then constructed that enforces near-equality of average word length and matches the 90th percentile of per-sentence max-word length via bounded swaps.

All 3 sets are filtered by BERT token length $\ge\!10$.

\subsection{Abstractness lexicon (short terms).}
This paper builds a lexicon of short terms consisting of at most two words and split into the following abstractness classes: 
(1) direct references to physical objects/actions/properties; 
(2) concepts with direct physical embodiment; 
(3) indirectly embodied or process-defined concepts; 
(4) abstract concepts with no clear relationship to the physical world. 
OpenAI o3 is instructed to generate 50-100 terms for each class provided the aforementioned criteria, with balanced variance in symbol length between the terms within each class; the generated dataset is then checked by a human, and inappropriate entries are removed. The final set (accessible at \emph{\url{https://github.com/MaxKho/SCAPE/blob/main/data/Abstractness.csv}}) consists of 236 terms in total. Each term is scored by SCAPE-T as a standalone input.

\subsection{Evaluation.}
Given score dictionary $s(\cdot)$, pairwise accuracy computes 
\begin{equation}
    \frac{1}{|S||C|}\!\sum_{x\in C}\!\sum_{y\in S}\!\big[\mathbbm{1}\{s(x)\!>\!s(y)\}+\tfrac{1}{2}\mathbbm{1}\{s(x)\!=\!s(y)\}\big],
\end{equation}
where $C$ is the complex group and $S$ the simple group. For the abstractness set, $C$ is Category 4 and $S$ is Categories 1\&2; all cross-category pairs are compared (ties $=\frac{1}{2}$). Category 3 was excluded from evaluation for clearer contrasts. For all other sets, pairwise accuracy is computed directly.

\subsection{Hyperparameter tuning.}
All hyperparameters (\emph{Table \ref{tab:hyperparams}}) are tuned once on the \emph{token-length-controlled} development set by maximising pairwise accuracy (ties $=\frac{1}{2}$) between complex and simple groups. A compact $(1+\lambda)$ evolutionary strategy with elitist selection is used:
let $\mathbf{p}_t\!\in\!\Pi$ be the current parameter vector (bounds $\Pi$ given in code). At iteration $t$, generate $\lambda$ offspring:
\begin{equation}
\mathbf{p}_t^{(i)}=\mathrm{clip}_\Pi\!\big(\mathbf{p}_t+\boldsymbol\varepsilon^{(i)}\big),
\quad
\varepsilon^{(i)}_d \sim \mathcal{N}\!\big(0,\,[\sigma_\varepsilon\, (u_d\!-\!l_d)]^2\big),
\end{equation}
where $[l_d,u_d]$ is the bound for dimension $d$ and $\sigma_\varepsilon$ is the mutation rate. Evaluate the objective $J(\mathbf{p})$ (pairwise accuracy) and set
$\mathbf{p}_{t+1}=\arg\max\{J(\mathbf{p}_t), J(\mathbf{p}_t^{(1)}),\dots,J(\mathbf{p}_t^{(\lambda)})\}$.
This paper uses $N_{t}\!=\!100$ iterations, $\lambda\!=\!10$, seed $=1$, $\sigma_\varepsilon\!=\!0.3$, and clip to bounds per dimension; the full hyperparameter bounds used are detailed in Appendix \ref{app:hyper}.

The selected settings are then frozen and used for all test sets and the abstractness lexicon. No test data is used for tuning. The following hyperparameter configuration was found by the evolutionary search (symbols as in \emph{Table~\ref{tab:hyperparams}}):
\[
\begin{aligned}
\ \alpha&=0.35,\ \beta=3.25,\ \gamma=0.40,\ \delta=10.0, \zeta=0.55,\\ \  \kappa&=0.10,\ \tau=0.90,\ \eta=1.00,\ \mu=6.75,\ \sigma=1.30.
\end{aligned}
\]

\section{Results}
\label{sec:results}

\subsection{Primary Results}
SCAPE-T is compared against:
\begin{enumerate}[label=(\roman*), leftmargin=2.2em]
\item Flesch–Kincaid readability (negated): negative Flesch reading ease from \texttt{textstat}.
\item Token length: number of BPE tokens from the GPT-2 tokeniser.
\item Type–token ratio (TTR): ratio of unique words to all words in the text.
\item Zipf frequency (negated mean): negative mean Zipf frequency using \texttt{wordfreq}, where
$\text{Zipf frequency}(w)=\log_{10}(\text{frequency of }w \text{ per }10^9 \text{ tokens})$.
\item GPT-2 perplexity: HuggingFace GPT-2 (\texttt{GPT2\-LMHead\-Model}) with default weights.
\item MDL bits/byte: total negative log-likelihood from GPT-2 converted to bits and normalised by UTF-8 bytes.
\item Lexicon complexity: mean lexical complexity rating per token from the DHH dataset introduced in \cite{Alonzo-TSAR-2022}.
\item Syntactic baseline: spaCy \texttt{en\_core\_web\_sm} parser for mean dependency-tree depth.
\end{enumerate}

\noindent All methods use the same sentences; bootstrap CIs are nonparametric, pair-level resampling percentile intervals given 100 resamples.

\begin{table}[!b]
\centering
\small
\setlength{\tabcolsep}{8pt}
\caption{Pairwise accuracy (\%).}
\label{tab:main-results}
\vspace{5pt}
\renewcommand{\arraystretch}{1.2}
\begin{tabular}{lccc}
\toprule
\textbf{Metric} & \textbf{WLC Test} & \textbf{Test} & \textbf{Abstractness} \\
\midrule
SCAPE-T                    & \textbf{78.5} & 85.1 & \textbf{82.3} \\
Lexicon complexity         & 65.8 & \textbf{94.3} & 74.4 \\
Flesch–Kincaid             & 59.5 & 93.1 & 78.9 \\
Type–Token ratio           & 56.4 & 61.7 & 50.0 \\
MDL (bits/byte)            & 52.0 & 34.1 & 26.7 \\
Zipf frequency             & 47.9 & 85.4 & 54.6 \\
GPT-2 perplexity           & 45.4 & 58.4 & 32.3 \\
Token length               & 39.2 & 49.6 & 51.7 \\
Mean dep-tree depth        & 36.9 & 45.5 & 50.6 \\
\bottomrule
\end{tabular}
\end{table}

On WLC Test (\emph{word-length–controlled} test set), SCAPE-T attains \textbf{78.5\%} pairwise accuracy with a 95\% bootstrap CI of 76.1–80.4, \textbf{+12.7pp} over the strongest baseline (DHH lexicon complexity, \(65.8\%\)); Flesch–Kincaid is \(59.5\%\), and the remaining baselines are \(\le 56.4\%\) (Table~\ref{tab:main-results}). On the uncontrolled test set, readability/lexical heuristics dominate (lexicon \(94.3\%\), Flesch–Kincaid \(93.1\%\), Zipf \(85.4\%\)); SCAPE-T reaches \(85.1\%\), while the remaining baselines are \(\le 61.4\%\).

On the (\emph{token-length–controlled}) development set, SCAPE-T scores \textbf{87.1\%} (bootstrap 95\% CI: 0.81–0.91, \(N{=}40{,}000\) pairwise comparisons). 

On the short-terms \emph{abstractness} dataset, SCAPE-T leads with \textbf{82.3\%}, ahead of Flesch–Kincaid (\(78.9\%\)) and the lexicon baseline (\(74.4\%\)); other baselines are \(\le 54.6\%\).

Additionally, as a test of linguistic bias, sentences from \emph{Le Petit Prince} (Project Gutenberg Australia) were contrasted against the development \emph{complex} group; the French children’s sentences were scored higher in \textbf{78\%} of pairs. A similar tendency was observed for single dialectal terms from Scottish English, which received disproportionately high scores.

\subsection{Ablations}
\label{sec:ablations}
Ablation analysis was performed on both the submetrics and the layers. Single-factor ablations were used for both, which zero one component at a time and re-evaluate pairwise accuracy with the same protocol. Both the development and the abstractness set were used for submetric ablations, while only the development set was used for layer ablations.

\paragraph{Submetric ablations}

CLS focus ablation demonstrated substantial drops on both sets, indicating it is likely the most informative submetric. Layer weighting ablation also demonstrated substantial drops on both sets. Self-focus ablations demonstrated smaller, but still statistically significant drops ($\Delta$ 95\% bootstrap CI: -5.67 to -3.42). Head redundancy ablations did not reveal a statistically significant contribution on the development set (full sentences), but did reveal a notable contribution on the abstractness set (short terms). 

\begin{table}[!b]
\centering
\small
\caption{Submetric ablations on development set.}
\label{tab:abl-dev}
\vspace{5pt}
\renewcommand{\arraystretch}{1.15}
\begin{tabular}{@{}l@{}r r@{}}
\toprule
\textbf{Ablation} & \textbf{Pairwise Acc.\ (\%)} & \(\boldsymbol{\Delta}\) \textbf{from base (pp)} \\
\midrule
Baseline                              & 88.80 & \(0.00\) \\
No head redundancy (\(\alpha{=}0\))   & 88.76 & \(-0.04\) \\
No self-focus penalty (\(\delta{=}0\))& 84.20 & \(-4.60\) \\
Uniform layer weights (\(\eta{=}0\))  & 58.62 & \(-30.18\) \\
No CLS focus (\(\beta{=}0\))          & 57.42 & \(-31.38\) \\
\bottomrule
\end{tabular}
\end{table}

\begin{table}[!b]
\centering
\small
\caption{Submetric ablations on abstractness set.}
\label{tab:abl-stt}
\vspace{7pt}
\renewcommand{\arraystretch}{1.15}
\begin{tabular}{@{}l@{}r r@{}}
\toprule
\textbf{Ablation} & \textbf{Pairwise Acc.\ (\%)} & \(\boldsymbol{\Delta}\) \textbf{from base (pp)} \\
\midrule
Baseline                              & 82.32 & \(0.00\) \\
No self-focus penalty (\(\delta{=}0\))& 81.04 & \(-1.29\) \\
No head redundancy (\(\alpha{=}0\))   & 80.25 & \(-2.08\) \\
Uniform layer weights (\(\eta{=}0\))  & 76.29 & \(-6.04\) \\
No CLS focus (\(\beta{=}0\))          & 64.60 & \(-17.72\) \\
\bottomrule
\end{tabular}
\end{table}

\paragraph{Per-layer ablations}
Layer 2 and layer 10 ablations demonstrated substantial drops; layer 1 and layer 3 ablations demonstrated minor gains. At all other layers, no notable changes were observed.

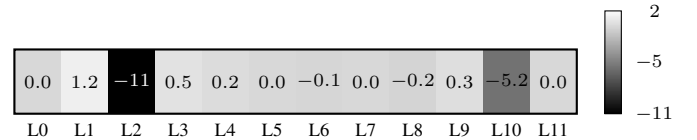
\begin{figure}[!b]
\centering
\begin{tikzpicture}[x=0.62cm,y=0.85cm]
  \def\minv{-11}   
  \def\maxv{  2}   
  \def\thr {0.60}  

  \def\data{%
    {0/L0/0.0},
    {1/L1/1.2},
    {2/L2/-11.0},
    {3/L3/0.5},
    {4/L4/0.2},
    {5/L5/0.0},
    {6/L6/-0.1},
    {7/L7/0.0},
    {8/L8/-0.2},
    {9/L9/0.3},
    {10/L10/-5.2},
    {11/L11/0.0}%
  }

  \foreach \x/\lab/\val in \data {
    \pgfmathsetmacro{\norm}{(\val-\minv)/(\maxv-\minv)} 
    \pgfmathsetmacro{\dark}{1 - \norm}                  
    \pgfmathsetmacro{\pct}{100*\dark}                   
    \fill[black!\pct] (\x,0) rectangle ++(1,1);

    \pgfmathparse{\dark>\thr?1:0}
    \ifnum\pgfmathresult=1 \def\tc{white}\else\def\tc{black}\fi

    \node[font=\scriptsize\bfseries, text=\tc] at (\x+0.5,0.5)
          {{\ifnum\pdfstrcmp{\val}{-11.0}=0
              \num[round-mode=places,round-precision=0]{\val}
            \else
              \num[round-mode=places,round-precision=1]{\val}
            \fi}};

    \node[font=\scriptsize, text=black] at (\x+0.5,-0.25) {\lab};
  }

  \draw[thick] (0,0) rectangle (12,1);

    \pgfmathsetmacro{\midv}{(\minv+\maxv)/2} 

  \begin{scope}[shift={(12.6,0)}]
    \shade[top color=white, bottom color=black] (0,0) rectangle (0.35,1.6);
    \draw (0,0) rectangle (0.35,1.6);
    \node[anchor=west, font=\scriptsize] at (0.45,0.00)
      {{\sisetup{round-mode=places,round-precision=0}\num{\minv}}};
    \node[anchor=west, font=\scriptsize] at (0.45,0.80)
      {{\sisetup{round-mode=places,round-precision=0}\num{\midv}}};
    \node[anchor=west, font=\scriptsize] at (0.75,1.60)
      {{\sisetup{round-mode=places,round-precision=0}\num{\maxv}}};
  \end{scope}
\end{tikzpicture}
\caption{Per-layer single-removal accuracy drop on development (\(\Delta\) accuracy in pp). Baseline accuracy: 88.8\%.}
\label{fig:layer-heat}
\end{figure}

\section{Discussion}
\label{sec:analysis}

\textbf{Overall patterns.}
On the uncontrolled test set, readability and lexical baselines obtain the highest scores, with SCAPE-T also competitive. Under word-length control (WLC), those baselines degrade sharply, while SCAPE-T remains strong, showing the highest performance of the tested baselines (Table~\ref{tab:main-results}); this corroborates the original proposition that they mostly track word length/frequency, not semantic integration. Similarly, on one–two-word items with minimal surface cues (the abstractness set), SCAPE-T leads as well, supporting sensitivity to conceptual load beyond readability. SCAPE-T's leading performance on the WLC and abstractness datasets indicates that SCAPE-T's comparatively lower accuracy than the leading baselines on the uncontrolled test set can reasonably be attributed, at least in part, to reduced sensitivity to predictive surface cues. On the other hand, GPT-2 perplexity and bits/byte underperform on WLC and abstractness, reflecting sensitivity to distributional mismatch and compressibility.

\textbf{Domain transfer.} Hyperparameters tuned on the token-length–controlled Hegel/Kant vs.\ MCTest contrast transfer to Wikipedia vs.\ children’s stories (encyclopedic technical prose vs.\ early-reader fiction), constituting evidence for genre and style generalisation.

\textbf{Linguistic bias.} The linguistic bias results are consistent with the expectation (mentioned in \S\ref{sec:intro}) that disproportionately high scores might indicate semantic integration blindspots: because SCAPE-T reads English BERT Base attention, extra context is needed to (partially) decode non-standard/non-English inputs, inflating scores. Encoders trained on multilingual data are therefore advisable for such data; see \S\ref{sec:limitations}.

\textbf{Layer ablations.} The result that layer 2 and layer 10 are the most informative by a considerable margin is unexpected. A potential explanation is that inverse self-focus (token interdependence) depends on early syntactic scaffolding (L2) and late semantic integration (L10); intervening layers contribute less specifically to this estimator. Further analysis is left to future work. The phenomenon is also discussed in \S\ref{sec:limitations}.

\textbf{Takeaways.}
(i) SCAPE-T isolates semantic integration under length control, and transfers across genres; (ii) CLS focus and layer weighting drive performance; inverse-weighted self-focus adds local-context sensitivity; (iii) head redundancy matters more on short inputs; (iv) layer ablations suggest early+late reliance for interdependence -- an open limitation to probe.

\section{Limitations}
\label{sec:limitations}
While SCAPE-T demonstrates stable performance and outperforms tested baselines under word-length control, a number of limitations remain. The metric reads raw encoder attention and should be treated as a diagnostic proxy for semantic integration rather than as a causal account of model reasoning. The operational definition is instantiated using a single frozen situation model -- BERT Base with an English tokeniser -- and to a CLS-style aggregator; portability to other encoders, sizes, tokenisers, architectures without a CLS token, and multilingual settings has not yet been established, so absolute scores are not guaranteed to be comparable across models. Moreover, SCAPE-T uses parametric knowledge as a proxy for true world state; because this knowledge is conditional on the pretraining corpus, it inherits its coverage gaps and biases \cite{nangia-etal-2020-crows}. Since these biases tend to affect conceptual knowledge rather than basic empirical facts, their effect on SCAPE-T’s estimates is unlikely to dominate, but requires investigation.

Per-layer ablations concentrate importance at specific layers, namely layers 2 and 10; while ablations also reveal that emphasising mid-to-late layers appreciably improves performance -- consistent with the premise that the semantic complexity signal is strongest in these layers -- the near-exclusive importance of layers 10 and especially 2 requires explanation.

Evaluation relies on proxy contrasts that are widely accepted but not gold standards for semantic content. Length control suppresses surface cues yet can distort natural length distributions and may underweight cases where length covaries with meaning density. The current scope is sentence-level; discourse phenomena such as coreference chains, rhetorical structure, and long-range coherence are only indirectly reflected when multi-sentence inputs are averaged. Preliminary observations indicate inflated scores for non-English or dialectal inputs processed with an English encoder, underscoring the need for multilingual validation.

Hyperparameters are tuned once on a single development contrast; although test domains are held out, some overfitting risk remains, and stability under alternative development splits is not exhaustively characterised; the relatively high number of hyperparameters exacerbates this risk. The head-redundancy term has limited effect on full sentences but a larger effect on short terms, suggesting length-dependent utility and design choices that could be improved. Computing and aggregating full attention maps across layers and heads incurs higher cost than readability heuristics, and the inverse self-focus transform improves accuracy at the expense of simple interpretability of raw magnitudes. Pairwise comparisons are therefore more reliable than absolute values. Given the observed 78.5\% pairwise accuracy on the \emph{word-length-controlled} test set, the metric is best deployed to capture corpus-level tendencies; it is not yet an error-proof indicator of semantic complexity for individual texts. Pairwise or group-level comparisons therefore remain the most defensible use case.

As mentioned in  \S\ref{subsec:scope}, SCAPE-T is extensible via complementary internal signals that theoretically satisfy a monotonic relationship with semantic complexity; therefore, the full space of such signals remains untested.

\section{Conclusion}
\label{sec:conclusion}
This paper formalises semantic complexity as the description length of a text's semantic content, and operationalises it with SCAPE-T, which aggregates three attention-derived correlates -- CLS focus, inverse-weighted self-focus, and inter-head redundancy -- via semantics-weighted layer aggregation. Under word-length control, SCAPE-T outperforms readability, lexical, syntactic, language-model, and compressor baselines on held-out domains, and leads on a style- and domain-neutral abstractness set; ablations indicate that each submetric and the aggregation scheme contribute materially. Taken together, the results provide a strong proof of concept that encoder attention patterns can be used to systematically gauge semantic complexity and lay practical groundwork for future research on semantic complexity estimation.

Promising directions include evaluating portability across models (such as Modern BERT, encoder-decoder models, and multilingual models such as XLM-R); extending from sentences to discourse-aware variants that incorporate coreference and rhetorical structure; integrating complementary internal signals such as attention rollout, inverse CLS self-focus (\texttt{[CLS]}→\texttt{[CLS]}), special handling of the \texttt{[SEP]} token, and gradient- or relevance-based attributions within the same monotonic framework; refining capacity-use terms beyond cosine-thresholded redundancy and reducing length dependence; constructing challenge sets targeted at semantic content to strengthen construct validity; and improving calibration so that absolute scores are comparable across models and domains without relying solely on pairwise evaluation.

\bibliographystyle{IEEEtran}
\bibliography{references}

\clearpage

\appendices
\section{Abbreviations, acronyms, and terminology}\label{app:abbr}

\small
\begin{table}[H]
\centering
\caption{Abbreviations, technical acronyms, and shorthand.}
\vspace{6pt}
\renewcommand{\arraystretch}{1.12}
\setlength{\tabcolsep}{6pt}
\begin{tabular}{@{}ll@{}}
\toprule
\textbf{Abbrev.} & \textbf{Meaning} \\
\midrule
\texttt{NLP}     & Natural Language Processing \\
\texttt{[CLS]}   & BERT classification token \\
\texttt{[SEP]}   & BERT separator token \\
\texttt{LLM}     & Large Language Model \\
\texttt{MDL}     & Minimum Description Length \\
\texttt{LSA}     & Latent Semantic Analysis \\
\texttt{RST}     & Rhetorical Structure Theory \\
\texttt{WLC}     & Word-length-controlled (test split) \\
\texttt{TTR}     & Type--Token Ratio \\
\texttt{BPE}     & Byte Pair Encoding \\
\texttt{CSV}     & Comma-Separated Values \\
\texttt{CI}      & Confidence Interval \\
\texttt{pp}      & Percentage points \\
\texttt{N}       & Sample size (count of items/pairs) \\
\bottomrule
\end{tabular}
\end{table}

\section{Preprocessing Details}\label{app:preproc}

\subsection{Overview}
The pipeline performs: (1) text-level normalisation; (2) tokenisation and sentence grouping; (3) per-sentence cleaning with rejections and repairs; (4) content-word projection (drop basic stopwords as well as punctuation except apostrophes); (5) per-sentence scoring and mean aggregation to document level.

\subsection{Text-level normalisation}
\paragraph{Ellipses}
Ellipses (``...'') do not provide much semantic information, and are therefore replaced context-sensitively as follows: if the next non-space character is uppercase, map to ``. ''; if lowercase, map to `` ''; otherwise leave as ``...''.

\paragraph{Abbreviation / citation expansion}
To prevent spurious sentence breaks and standardise markers, the following replacements are applied once to common abbreviations, left to right:

\begin{itemize}[leftmargin=*,nosep]
  \item \textbf{Honorifics and titles}
  \begin{itemize}[noitemsep,topsep=0pt,leftmargin=1.5em]
    \item \texttt{Dr.} $\rightarrow$ \texttt{Doctor}
    \item \texttt{Prof.} $\rightarrow$ \texttt{Professor}
    \item \texttt{Sr.} $\rightarrow$ \texttt{Senior}
    \item \texttt{Jr.} $\rightarrow$ \texttt{Junior}
    \item \texttt{St.} $\rightarrow$ \texttt{Saint}
    \item \texttt{Mr.}, \texttt{Mrs.}, \texttt{Ms.}: drop the period and keep a trailing space (e.g., \texttt{Mr.} $\rightarrow$ \texttt{Mr }).
  \end{itemize}

  \item \textbf{Latinisms}
  \begin{itemize}[noitemsep,topsep=0pt,leftmargin=1.5em]
    \item \texttt{i.e.} $\rightarrow$ \texttt{that is, }
    \item \texttt{e.g.} $\rightarrow$ \texttt{for example }
    \item \texttt{viz.} $\rightarrow$ \texttt{namely }
    \item \texttt{cf.} $\rightarrow$ \texttt{cf } (period removed, space kept)
  \end{itemize}

  \item \textbf{Bibliographic shorthands}
  \begin{itemize}[noitemsep,topsep=0pt,leftmargin=1.5em]
    \item \texttt{p.} $\rightarrow$ \texttt{page } \quad / \quad \texttt{P.} $\rightarrow$ \texttt{Page }
    \item \texttt{pp.} $\rightarrow$ \texttt{pages }
    \item \texttt{l.} $\rightarrow$ \texttt{line }
    \item \texttt{ch.} or \texttt{chap.} $\rightarrow$ \texttt{chapter }
    \item \texttt{ed.} $\rightarrow$ \texttt{edition } \quad / \quad \texttt{edd.} $\rightarrow$ \texttt{editions }
    \item \texttt{vol.} or \texttt{Vol.} $\rightarrow$ \texttt{volume }
    \item \texttt{pref.} $\rightarrow$ \texttt{preface }
  \end{itemize}

  \item \textbf{\texttt{etc.} and \texttt{\&c.}}
  \begin{itemize}[noitemsep,topsep=0pt,leftmargin=1.5em]
    \item Standardise both to \texttt{and so forth}.
    \item If a capitalised word follows immediately (e.g., \texttt{etc. Quantum}), ensure there is a period and a space before the capitalised word: \texttt{and so forth . Quantum}.
  \end{itemize}

  \item \textbf{Roman numerals}
  \begin{itemize}[noitemsep,topsep=0pt,leftmargin=1.5em]
    \item If a Roman numeral (e.g., \texttt{IV}, \texttt{xvi}) is followed by a period, drop the period and leave a trailing space (e.g., \texttt{xvi.} $\rightarrow$ \texttt{xvi }).
  \end{itemize}

  \item \textbf{Consecutive punctuation marks}
  \begin{itemize}[noitemsep,topsep=0pt,leftmargin=1.5em]
    \item Collapse any sequence of repeated punctuation (\texttt{. , ; : ? !}) to a single instance (e.g., \texttt{?!?} $\rightarrow$ \texttt{?}, \texttt{,,} $\rightarrow$ \texttt{,}).
  \end{itemize}
\end{itemize}

\subsection{Segmentation and sentence grouping}
\paragraph{Segmentation}
Each text is segmented into (i) maximal non-space spans that exclude \texttt{\{.,,,:;?!\}} -- called \emph{words} in this paper -- and (ii) single punctuation marks from that set -- called \emph{punctuation}. (Regex used: two-alternative pattern equivalent to \textit{non-space, non-punct} or \textit{single punct}.)

\paragraph{Sentence grouping}
The set of segments is grouped into sentences as follows: tokens are scanned left-to-right; on encountering a delimiter (\texttt{., !, ?}), the current sentence is ended, and the delimiter is attached to it. A trailing fragment without a delimiter is kept.

\subsection{Per-sentence cleaning}
Each raw sentence undergoes: hard rejections, repairs, a fixed-point filter, and final sanitation.

\paragraph{Hard rejections (drop the sentence if any triggers)}
\begin{enumerate}[leftmargin=*,nosep]
  \item If the sentence contains a span enclosed in single quotes (e.g., ’…’), treat it as direct speech and discard: direct speech often expresses a separate idea to the sentence it is contained in, making sentence-level semantics ambiguous.
  \item If the sentence contains obvious formatting/artefact markers, namely \texttt{•}, \texttt{”}, \texttt{"}, \texttt{§}, an isolated apostrophe, or an isolated \texttt{i} token, treat the sequence as formatting and discard.
  \item If the sentence begins with a number followed by a colon (e.g., “12:” or “3 :”), treat it as a timestamp/list marker and discard.
\end{enumerate}

\paragraph{Repairs (single pass, in this order)}
\begin{enumerate}[leftmargin=*,nosep]
  \item Remove a trailing stray \verb|_|, \texttt{]}, or \texttt{)} at end of sentence.
  \item Remove a leading \texttt{[} or \texttt{(} at start of sentence.
  \item Delete stray inline \texttt{[} or \texttt{(} (unmatched openers).
  \item Strip a leading one of \texttt{, : ;}.
  \item Normalise em-dash spacing around \texttt{--}: ensure “[letter]\texttt{--}” becomes “[letter]\ \texttt{--}” and “\texttt{--}[letter]” becomes “\texttt{--}\ [letter]”.
  \item Collapse consecutive punctuation marks (\texttt{, : ; . ? !}) to a single mark (e.g., \texttt{?!?} → \texttt{?}).
  \item Collapse consecutive runs of \texttt{-} to a single \texttt{-}.
  \item If the sentence starts with \texttt{Page} or \texttt{Pages} followed by a number (e.g., \texttt{Page 12}), remove that prefix.
  \item Remove empty single-quote pairs at the start (e.g., \texttt{'\ ' '} with optional spaces).
  \item Strip a leading number at the start.
  \item Remove leading asterisk bullets or runs of \texttt{*}.
\end{enumerate}

\paragraph{Fixed-point filter (repeat this sequence until no further change)}
\begin{enumerate}[leftmargin=*,nosep]
  \item If the sentence starts with an unmatched \texttt{' } or \texttt{"}, drop that leading quote.
  \item Strip any leading digits.
  \item Strip one leading symbol if it is one of \texttt{, : ; . -}.
  \item If the sentence begins with a Roman-numeral run from \texttt{\{i,v,x,I,V,X\}} \emph{or} with \texttt{page}/\texttt{Page}, optionally followed by \texttt{page}, then a number, remove that whole prefix (e.g., \texttt{iv 23}, \texttt{Page 5}, \texttt{Page page 5}).
  \item If it begins with \texttt{line <digits> from bottom} or \texttt{line <digits> from top}, remove that whole prefix.
  \item If it ends with \texttt{,} + optional spaces + (Roman-numeral run \emph{or} \texttt{page}) + spaces + digits + optional spaces + optional \texttt{.}, remove that trailing reference.
  \item Remove any remaining leading digits (if produced by earlier edits).
  \item Remove the odd leading sequence \texttt{',}.
  \item Remove a leading run of two or more Roman-numeral letters from \texttt{\{i,v,x,I,V,X\}}.
\end{enumerate}

\paragraph{Final cleaning}
\begin{enumerate}
    \item Collapse consecutive spaces. 
    \item Remove leading and trailing spaces.
    \item If the sentence is non-empty and doesn't already end in a delimiter (\{., !, ?\}), append `` .''.
    \item Collapse any remaining punctuation runs using the same ``collapse consecutive punctuation marks'' rule as above. 
    \item If the sentence begins with a lowercase letter, treat as a sentence grouping error and discard.
\end{enumerate}

The cleaning procedure was motivated partially by universal tendencies and partially by Project Gutenberg idiosyncrasies. An alternative approach would have been to devise two separate cleaning algorithms, with one being applied to all SCAPE-T input and one applied only to Project Gutenberg texts; however, for simplicity, as well as to streamline the cleaning process for future researchers intending to evaluate SCAPE-T on Project Gutenberg texts, the two were combined into one preprocessing pipeline.

\subsection{Content-word projection (before scoring)}
From each cleaned sentence:
\begin{enumerate}
    \item remove ASCII punctuation \emph{except} apostrophes (so contractions are preserved)
    \item drop the stop words ``a'', ``an'', ``the'', ``that'', ``and'', ``very'', ``will'', ``just'', which carry little semantic information and tend to deflate complexity scores
    \item Sentences that become empty after (i)–(ii) should be discarded.
\end{enumerate}

\subsection{Aggregation to document level}
Given per-sentence scores \(g(r_k)\) for \(k=1,\dots,m\), the document score is the arithmetic mean
\[
S_{\mathrm{doc}}=\frac{1}{m}\sum_{k=1}^{m} g(r_k),
\]
with the document skipped if \(m=0\).

\subsection{Determinism}
All steps are functional (no sampling). The encoder is frozen and run in evaluation mode with attention outputs enabled, so preprocessing is deterministic given the input.

\section{Hyperparameter Bounds}\label{app:hyper}
The following hyperparameter bounds were used for the evolutionary search:

\begin{table}[H]
\centering
\small
\setlength{\tabcolsep}{10pt}
\caption{Hyperparameter bounds.}
\label{tab:hparam-bounds}
\vspace{5pt}
\renewcommand{\arraystretch}{1.15}
\begin{tabular}{lc}
\toprule
\textbf{Parameter} & \textbf{Search range} \\
\midrule
$\alpha$ & $[\num{0.0}, \num{3.0}]$ \\
$\beta$  & $[\num{0.0}, \num{3.0}]$ \\
$\gamma$ & $[\num{0.0}, \num{3.0}]$ \\
$\delta$ & $[\num{0.0}, \num{3.0}]$ \\
$\zeta$  & $[\num{0.0}, \num{1.0}]$ \\
$\kappa$ & $[\num[round-mode=places,round-precision=1]{0.1}, \num{5.0}]$ \\
$\tau$   & $[\num[round-mode=places,round-precision=1]{0.1}, \num[round-mode=places,round-precision=1]{0.9}]$ \\
$\eta$   & $[\num{0.0}, \num{1.0}]$ \\
$\mu$    & $[\num{3.0}, \num[round-mode=places,round-precision=1]{11.0}]$ \\
$\sigma$ & $[\num{1.0}, \num{4.0}]$ \\
\bottomrule
\end{tabular}
\end{table}

\end{document}